\documentclass[10pt,twocolumn,letterpaper]{article}

\usepackage{cvpr}
\usepackage{times}
\usepackage{epsfig}
\usepackage{graphicx}
\usepackage{amsmath}
\usepackage{amssymb}
\usepackage{diagbox}

\cvprfinalcopy 

\ifcvprfinal\fi
\begin{document}

\title{Learning Discriminative Features with Class Encoder}

\author{Hailin Shi,  Xiangyu Zhu,  Zhen Lei, Shengcai Liao,  Stan Z. Li \\
Center for Biometrics and Security Research \& National Laboratory of Pattern Recognition\\
Institute of Automation, Chinese Academy of Sciences\\
University of Chinese Academy of Science \\
{\tt\small \{hailin.shi, xiangyu.zhu, zlei, scliao, szli\}@nlpr.ia.ac.cn}
}


\maketitle
\thispagestyle{empty}

\begin{abstract}
   Deep neural networks usually benefit from unsupervised pre-training, e.g. auto-encoders.
    However, the classifier further needs supervised fine-tuning methods for good discrimination.
    Besides, due to the limits of full-connection, the application of auto-encoders is usually limited to small, well aligned images.
    In this paper, we incorporate the supervised information to propose a novel formulation, namely class-encoder, whose training objective is to reconstruct a sample from another one of which the labels are identical.
    Class-encoder aims to minimize the intra-class variations in the feature space, and to learn a good discriminative manifolds on a class scale.
    We impose the class-encoder as a constraint into the softmax for better supervised training, and extend the reconstruction on feature-level to tackle the parameter size issue and translation issue.
    The experiments show that the class-encoder helps to improve the performance on benchmarks of classification and face recognition.
    This could also be a promising direction for fast training of face recognition models.
\end{abstract}


\section{Introduction}

In recent years, many learning algorithms, \eg Restricted Boltzmann Machine (RBM) \cite{hinton2006reducing} and auto-encoder (AE) \cite{bengio2007greedy}, proposed to pre-train the neural network by auto-reconstruction in a layer-wise way and achieved breakthroughs on training problems.
This sort of algorithms, to which we refer as \emph{reconstructive} methods, constitute an important subset of deep learning approaches nowadays.
More recently, along this direction, certain variants of AE, such as denoising auto-encoder (DAE) \cite{vincent2008extracting,vincent2010stacked} and contractive auto-encoder (CAE) \cite{rifai2011contractive}, referred to as regularized AEs \cite{alain2012regularized}, are proposed to estimate data-generating distribution on a local scale and learn compact low-dimensional manifolds, in which better discrimination power can be expected.

On the other hand, convolutional neural networks (CNN) \cite{lecun1998gradient} is also a widely-used approach of deep learning towards computer vision.
In recent years, the computational resources have been massively improved by GPU implementations \cite{krizhevsky2012imagenet,jia2014caffe} and distributed computing clusters \cite{dean2012large}, and various large-scale data sets have been collected to satisfy the training.
Due to these benefits, CNNs demonstrated the power of hierarchical representation by beating the hand-craft features, and won many contests in this field \cite{krizhevsky2012imagenet,sermanet2013overfeat,girshick2014rich,szegedy2014going}.

\textbf{Problems.}
Firstly, RBM, AE and their variants are unsupervised methods. To bring about good discrimination, the classifier needs supervised training. In other words, good representation from reconstruction does not guarantee good classification \cite{alain2012regularized}. This suggests to find an objective with both reconstructive and discriminative aspects to improve the training.

Secondly, the auto-encoders are not robust to image translation; in addition, they often keep a large number of parameters that increase explosively according to the data size. As a result, the application of AE is usually limited to small, well aligned images.

\textbf{Contribution.}
Firstly, we propose a supervised reconstructive model, referred to as \emph{class-encoder}, whose objective is the reconstruction of one sample from another within the same class.
The model minimizes the intra-class variations and learns compact low-dimensional manifolds on a class scale.
Although class-encoder method is similar to AE, its application is not in the pre-training.
Class-encoder is directly used in the supervised training of network, as it is a supervised method.
We further imposed the class-encoder as a constraint into the softmax classifier (namely Class-Encoding Classifier, CEC) and achieve better performance than the pure softmax.

Secondly, we propose a deep hybrid neural network that combines the CNN and the CEC, so to let them benefit from each other.
The convolutional layers extract features from data at the bottom level, and the CEC is disposed at the top level.
Different from former reconstructive models which directly reconstructs data, in this framework, the intra-class reconstruction is performed on the feature-level.
So, the CEC is robust to translation due to the CNN, and CNN has better generalization thanks to the CEC.
Besides, the size of fully-connected (FC) layer and its parameter number are limited in an acceptable range, because the reconstructive target is not images but feature vectors.
We use this network to learn robust and discriminative features for face recognition.

\section{Related work}

\textbf{Regularized auto-encoders.} DAE and CAE locally estimates data-generating distribution and captures local manifold structure. Their pre-training is based on unsupervised method. By contrast, class-encoder extends them to a supervised style.

\textbf{FIP feature.} Zhu \etal \cite{zhu2013deep} proposed to learn face identity-preserving (FIP) features through recovering frontal face images from other views.
Another work \cite{zhu2014recover} employed a similar method which trained multiple deep networks on the facial components of recovered frontal face.
Comparing with class-encoder, their training objective is strictly fixed by canonical view. Therefore, the selection of canonical view is indispensable.
Besides, their reconstruction is performed on data-level, not feature-level. Thus, the performance is very limited by data condition, \ie facial expression, image cropping (background interference), alignment \etc
The feature-level reconstruction of class-encoder is crucial for the elimination of nuisance factors.

\section{The proposed method}
In this section, we begin with class-encoder. Then, we introduce the CEC model. Finally, we describe the Deep CEC.

\subsection{Class-encoder}
\begin{figure}[!htb]
  \centering
  \includegraphics[width=0.14\textwidth]{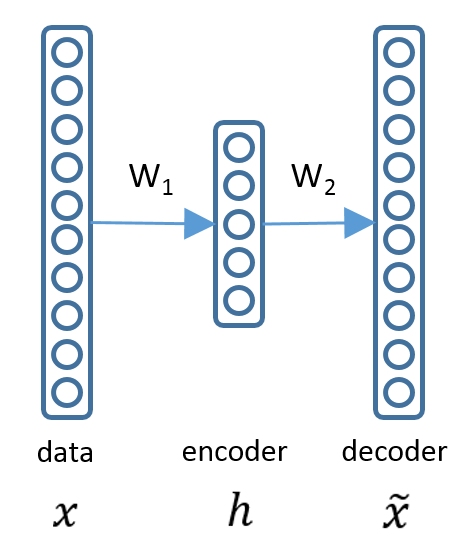}
  \caption{Class-encoder network with single layer of encoder and decoder.}
  \label{ce}
\end{figure}

Class-encoder and auto-encoder share the same architecture (Fig.~\ref{ce}) which includes an input layer, a hidden layer (encoder) and an output layer (decoder) of full-connection. The training objective is the main difference between class-encoder and auto-encoder. Auto-encoder aims to reconstruct a data sample from itself, while class-encoder performs the reconstruction of one sample from another one with the same label.

\textbf{Formulation.} Let $x$ be an input data, $h$ be the activation of the hidden layer, $\tilde{x}$ be the reconstruction, $W_1$ and $W_2$ be the weight matrices of the FC layers. $W_1$ and $W_2$ often take form of tied weights, \ie $W_1^T = W_2$, which is usually employed as an implicit regularization for preventing extremely large and small entries.
For the simplicity, we merge the bias term into the weight matrices in this paper. Then, the reconstruction $\tilde{x}$ is calculated as follows:
\begin{equation}\label{ce_reconstruction_1}
    h = f(W_1 x)
\end{equation}
\begin{equation}\label{ce_reconstruction_2}
    \tilde{x} = f(W_2 h) = f(W_2 f(W_1 x))
\end{equation}
where $f(\cdot)$ is the activation function.
To achieve intra-class reconstruction, let $\hat{x}$ be any data sample that has the same label with $x$. Therefore, the objective function of class-encoder is defined as
\begin{equation}\label{ce_objective_L2}
    \mathit{Cost}_{\mathit{ce}} = \frac{1}{2 N} \sum_{x \in \mathbf{X}} \sum_{\hat{x} \in \mathbf{S}_x} {\parallel \tilde{x} - \hat{x} \parallel}^2
\end{equation}
where $N$ denotes the total number of training data, $\mathbf{X}$ denotes the entire training data set, and $\mathbf{S}_x$ denotes the subset of the class in which $x$ is found. Supposing there are $\mathbf{C}$ classes in total, let $c = 1,2,...,\mathbf{C}$ be the class labels, and $S_c$ be the subset of $c^{th}$ class with size of $N_c$. Then, Eq.~\ref{ce_objective_L2} can be developed as follows:
\begin{align}\label{ce_objective_L2_inference}
    \mathit{Cost}_{\mathit{ce}} & = \frac{1}{2} \sum_{c=1}^{\mathbf{C}} \frac{1}{N_c} \sum_{x \in S_c} \sum_{\hat{x} \in S_c} {\parallel \tilde{x} - \hat{x} \parallel}^2 \nonumber  \\
    & = \frac{1}{2} \sum_{c=1}^{\mathbf{C}} \sum_{x \in S_c} \frac{1}{N_c} \sum_{\hat{x} \in S_c} ({\parallel \tilde{x} \parallel}^2 + {\parallel \hat{x} \parallel}^2 - 2{\tilde{x}}^T \hat{x}) \nonumber  \\
    & = \frac{1}{2} \sum_{c=1}^{\mathbf{C}} \sum_{x \in S_c} (\frac{1}{N_c} {\parallel \tilde{x} \parallel}^2 + \frac{1}{N_c} \sum_{\hat{x} \in S_c} {\parallel \hat{x} \parallel}^2 \nonumber  \\
     & \qquad \qquad \qquad  - 2{\tilde{x}}^T ( \frac{1}{N_c} \sum_{\hat{x} \in S_c} \hat{x})).
\end{align}

In Eq.~\ref{ce_objective_L2_inference}, the first term is regarded as a penalty of magnitude of the reconstruction; the second term is constant; the third term indicates that class-encoder's reconstruction $\tilde{x}$ is prone to have small angle with the mean vector of the corresponding class. Hence, class-encoder tends to maximize a cosine-similarity-like metric between the reconstructions and intra-class means.

It is a supervised learning task which implicitly minimizes the intra-class variation. The model a learns discriminative low-dimensional manifold on a class scale in the decoder space. Data points are projected into a dense distribution within each class, whose center is located at the intra-class mean. Considering Eq.~\ref{ce_reconstruction_1}, this intra-class convergency also takes place in the hidden layer $h$ (\ie encoder space). It will be proved empirically in the next section.

\subsection{CEC model}
To make use of the advantage that class-encoder minimizes the intra-class variation, we impose the class-encoder into the softmax classifier, and train the network with the intra-class reconstruction and softmax regression jointly, in order to potentiate the discrimination.

\begin{figure}[!htb]
  \centering
  \includegraphics[width=0.18\textwidth]{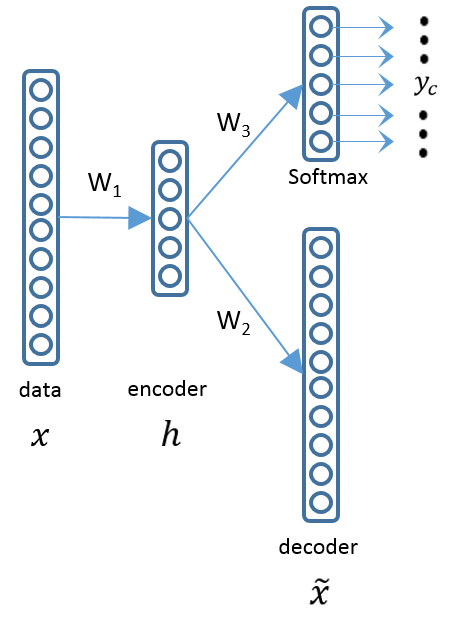}
  \caption{CEC model. We train class-encoder and softmax simultaneously. During the test of classification, we ignore the decoder and only take account of the output of softmax.}
  \label{cec}
\end{figure}

Fig.~\ref{cec} shows the model of CEC. The objective function is the weighted sum of each part,
\begin{equation}\label{cec_objective}
    \mathit{Cost}_{cec} = \mathit{Cost}_{\mathit{softmax}} + \lambda \mathit{Cost}_{\mathit{ce}}.
\end{equation}

The second term in Eq.~\ref{cec_objective} represents the weighted cost from class-encoder. It has the same definition with Eq.~\ref{ce_objective_L2}.
The cost of softmax is formulated as
\begin{equation}\label{softmax-cost}
    \mathit{Cost}_{\mathit{softmax}} = - \sum_{c=1}^{\mathbf{C}} \frac{1}{N_c} \sum_{x \in S_c} log \frac{exp(W_3^c h)}{\sum_{l=1}^{\mathbf{C}} exp(W_3^l h)},
\end{equation}
where $W_3^c$ and $W_3^l$ are the $c^{th}$ and $l^{th}$ row of $W_3$. The softmax outputs the probability that an input $h$ belongs to the $c^{th}$ class by computing the following equation
\begin{equation}\label{softmax-output}
    P(y=y_c\mid W_3, h) = \frac{exp(W_3^c h)}{\sum_{l=1}^{\mathbf{C}} exp(W_3^l h)},
\end{equation}
where $y_c \in \{1, \dots, \mathbf{C} \}  $ is the ground-truth class label of the $c^{th}$ class, and $y$ is the prediction decision.
Obviously, we expect this probability to be large for the correct prediction. This probability can be developed by the Bayesian rule
\begin{equation}\label{softmax-bayse}
    P(y=y_c\mid W_3, h) = P(h \mid y=y_c, W_3) \frac{P(y=y_c)} {P(h)}.
\end{equation}

We assume that the conditional probability $P(h \mid y=y_c, W_3)$ follows the Gaussian distribution,
\begin{equation}\label{cec_cond_proba_1}
    h \mid y=y_c, W_3 \quad \sim \quad \mathcal{N} (\mu, \sigma).
\end{equation}
It is also natural to assume the conditional of $h$ in the class $y_c$ follows the Gaussian distribution,
\begin{equation}\label{cec_cond_proba_2}
    h \mid y_c \quad \sim \quad \mathcal{N} (\mu_h, \sigma_h).
\end{equation}
From an optimized softmax classifier, we can find either $\mu$ = $\mu_h$ or the two mean vectors are very close. In addition, due to the effect of class-encoder, $\sigma_h$ is small. Thus, softmax has a very large probability to have $h$ close to $\mu$, which leads to a large value of $P(h \mid y=y_c, W_3)$ and so the output probability in Eq.~\ref{softmax-output}. In other words, the class-encoder improves the lower-bound of the likelihood of softmax. Sharper distribution $P(h \mid y_c)$ we sample from, more possibly we obtain large value of likelihood.

%
%
%

\subsection{Deep CEC and Feature-level Strategy}

\begin{figure}[!htb]
  \centering
  \includegraphics[width=0.45\textwidth]{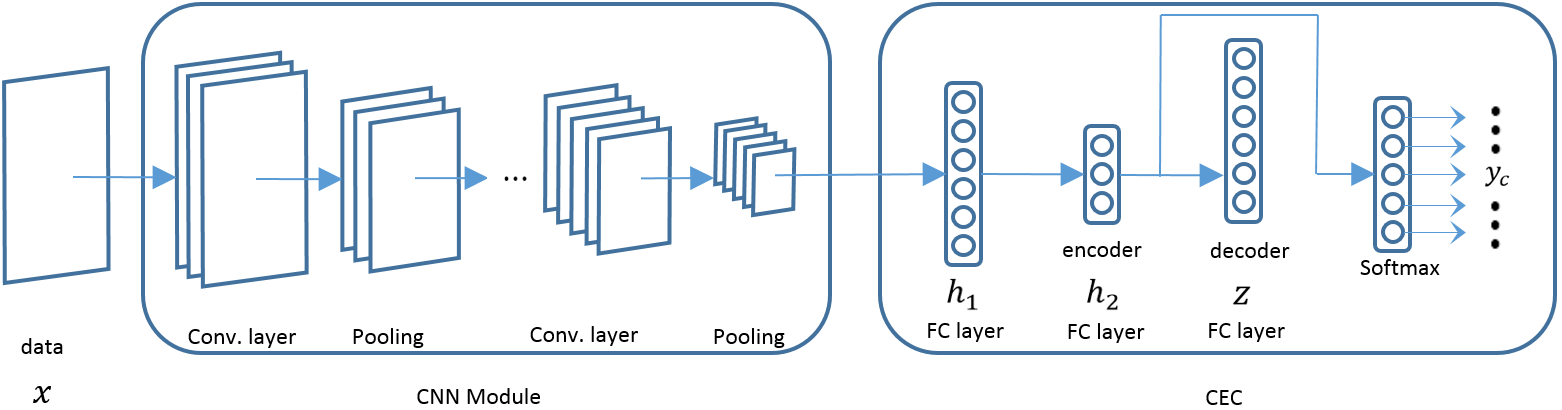}
  \caption{Overview of DCEC. The network is built by cascading the CNN module and the CEC.}
  \label{dcec}
\end{figure}

Deep CEC (DCEC) is built by cascading CNN module and the CEC (Fig.~\ref{dcec}).
Like conventional CNNs, the CNN module is composed by convolutional and max-pooling layers.

The CEC receives the features from the CNN module, and works like the above-mentioned CEC except that the decoder aims to reconstruct the feature rather than the raw data. Here, we notate the input data pair as \{$x'$, $x''$\}.
Let $h_1$, $h_2$, and $z$ be the activations of the first layer, encoder, and decoder in the CEC, respectively. The training objective is defined as
\begin{equation}\label{dcec_objective_L2}
    \mathit{Cost}_{\mathit{ce}}^{\mathit{feature}} = \frac{1}{2 N} \sum_{x' \in \mathbf{X}} \sum_{x'' \in \mathbf{S}_{x'}} {\parallel z' - h_1'' \parallel}^2.
\end{equation}
Note that $z'$ and $h_1''$ come from the input data pair \{$x'$, $x''$\}, not from a single sample. In the practical training, $x'$ and $x''$ are sampled from a class, and input to the DCEC in sequence, to compute $z'$ and $h_1''$, respectively.

Here, the objective of class-encoder is to reconstruct the features (\ie $h_1''$). We refer to this kind of reconstruction as \emph{feature-level}, in contrast to the data-level reconstruction. There are two reasons behind the feature-level reconstruction.

First, the images may contain not only the target object, but nuisance factors as well, such as background, facial expression, poses \etc Simply reconstructing the intra-class images will introduce substantial noise to the training, whereas the feature-level reconstruction can eliminate the nuisance factors, and preserve the discriminative factors in the feature space. This is because the input of CEC is no longer raw data, but features. Considering two input samples with the same label, their features' common part is the discriminative factors. It exists a large probability that the features have the accordance in discriminative factors, and the discrepancy in nuisance factors, since the nuisance factors are very likely different (\eg background in different images could seldom be the same). Therefore, by reconstruction from one to another in the same class, the proportion of nuisance factor is reduced in the feature space. From another point of view (\ie the previous interpretation of convergency), the intra-class features converge to the corresponding discriminative factor.

Second, the target object may present at different locations in images. Without alignment, the data-level reconstruction will introduce the noise too.
Owing to the CNN module, the extracted feature is robust to image translation, and so is the feature-level reconstruction.

The objective function of DCEC is the weighted sum of softmax and intra-class, feature-level reconstruction,
\begin{equation}\label{dcec_objective}
    \mathit{Cost}_{\mathit{dcec}} = \mathit{Cost}_{\mathit{softmax}} + \lambda \mathit{Cost}_{\mathit{ce}}^{\mathit{feature}}.
\end{equation}
By BP method, the CNN module and the CEC are trained simultaneously.

\section{Experiments}
In this section, we report the experiments of the proposed methods. We started with the pure class-encoder. Then, we extended the experiment to CEC. Finally, we applied DCEC to learn robust features for human face recognition.

\subsection{Inspection of class-encoder}
In this subsection, we trained a network of pure class-encoder, in order to give an intuitive show of class-encoder's ability of discrimination in the feature space.

\textbf{Data.} MNIST \cite{lecun1998gradient} is a general database of handwritten digits, containing a training set of 50,000 samples, a validation set of 10,000 samples, and a test set of 10,000 samples. The 10 digits own roughly equal number of samples.

\textbf{Setting.} To achieve good convergency, we built a 4-layer encoder and a symmetrical decoder. The number of nodes for encoder were 2000-1000-500-250, determined by referring to the architecture in Hinton \etal \cite{hinton2006reducing}.
Since the data had been well aligned and keep mono-black background, we let the reconstruction to be on data-level.
The network was randomly initialized.
We randomly selected 15,000 pairs for each digit. Each pair was fed to the network consequently to calculate the reconstruction cost.

\begin{figure}[!htb]
    \centering
    \includegraphics[width=0.19\textwidth]{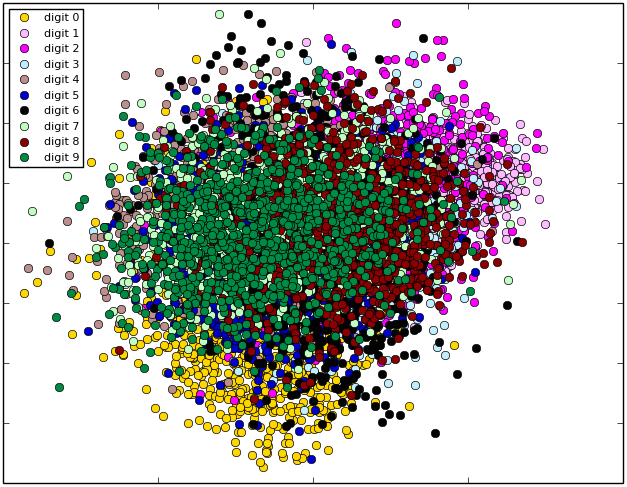}\hspace{0.6em}
    \includegraphics[width=0.19\textwidth]{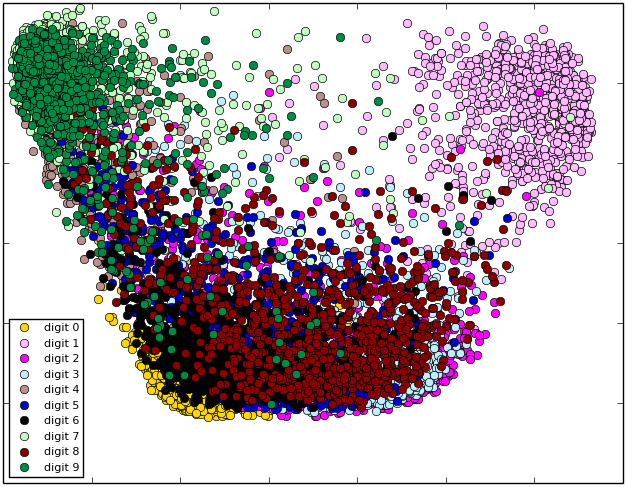}\\
    \includegraphics[width=0.19\textwidth]{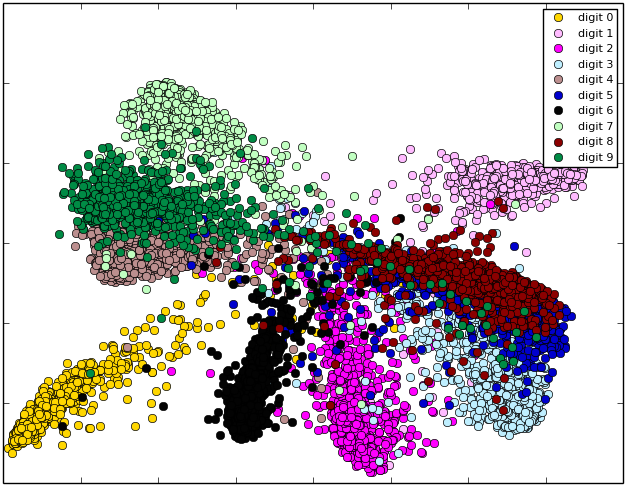}\hspace{0.6em}
    \includegraphics[width=0.19\textwidth]{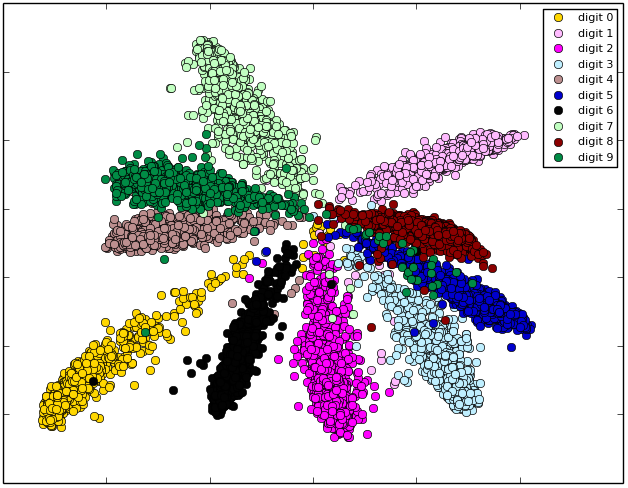}
    \caption{From left to right, top to bottom: scatters of the middle-layer activation of the class-encoder network along with the training epoch 0, 10, 50 and 200. We assign each digit a distinct color. Best view in color.}
    \label{fig_scatter_mlce}
\end{figure}

\begin{figure}[!htb]
    \centering
    \includegraphics[width=0.12\textwidth]{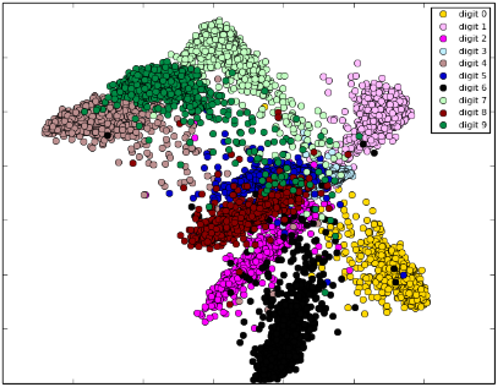}\hspace{0.01em}
    \includegraphics[width=0.12\textwidth]{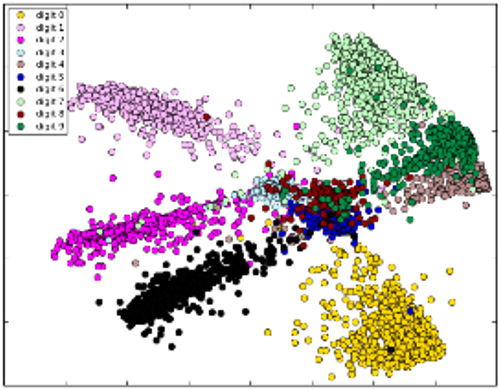}\hspace{0.01em}
    \includegraphics[width=0.12\textwidth]{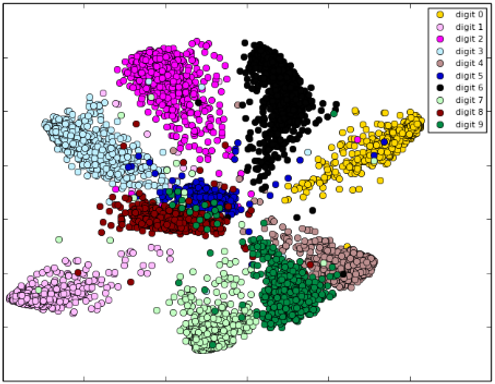}\\
    \caption{From left to right, the corresponding architectures of encoder are 200-200-200, 1500-1000-500, and 200-200-200-200, respectively. 
    }
    \label{fig_scatter_mlce_more_attempts}
\end{figure}

\textbf{Result.} The network was optimized by stochastic gradient descent (SGD) and BP method. We extracted the activation values of the middle layer (250-dimensional) and reduced its dimensionality to 2 by PCA. We show the scatters in Fig~\ref{fig_scatter_mlce}. Along with the training process, each class converged effectively. In Fig.~\ref{fig_scatter_mlce_more_attempts}, we show more attempts on different architectures. The scatters suggest that deeper and wider architectures give better results.

\subsection{CEC for classification}

In this subsection, we evaluated the CEC for classification.

\textbf{Data.} We evaluated the classification experiments on MNIST.

\textbf{Setting.} We chose the pure softmax as our baseline model.
We compared the pure softmax with CEC for classification task, in order to highlight the advantage of class-encoder.
Note that CEC drops into softmax when the weight $\lambda$ becomes 0 in Eq.~\ref{cec_objective}.

\begin{figure}[!htb]
  \centering
  \includegraphics[width=0.25\textwidth]{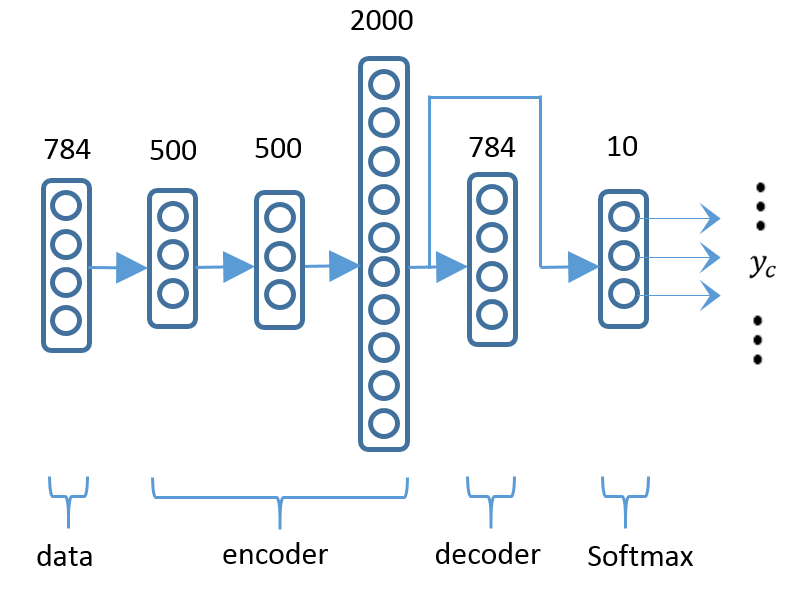}
  \caption{CEC with multi-layer encoder and single-layer decoder. The baseline was the same but without decoder.}
  \label{mlcec}
\end{figure}

Fig.~\ref{mlcec} shows the architecture of CEC.
The decoder was a single FC layer since, with a large number of experiments, we found that the one-layer decoder was most suitable for reconstruction.

For the diversity of experiment, we initialized the network in 3 different ways -- AE, DAE, and CAE. Then, we took the pre-trained networks for either CEC or softmax.

\textbf{Result.}
Table~\ref{tab_mlcec_error_rate} shows that our CEC outperforms the baselines on MNIST classification.
We found that the method of initialization (AE, DAE, or CAE) does not influence the CEC reaching better results.

It should be mentioned that the training error rate reached zero for all the models. Therefore, the class-encoder improved the classifier's generalization.

\begin{table}   \small
\newcommand{\tabincell}[2]{\begin{tabular}{@{}#1@{}}#2\end{tabular}}
\begin{center}
\begin{tabular}{|c|c|c|}
\hline
\diagbox{{Initialization}}{{Training}}
& {softmax} & \tabincell{c}{{CEC} \\ {(CE+softmax)}} \\
\hline
{AE} &   1.40$\pm$0.23 &   1.29$\pm$0.18 \\
\hline
{DAE} &   1.28$\pm$0.22 &   1.16$\pm$0.20 \\
\hline
{CAE} &   1.26$\pm$0.12 &   1.15$\pm$0.09 \\
\hline
\end{tabular}
\end{center}
\caption{Test error rates (in percentage) on MNIST.
 In each line, the baseline model was compared with CEC that initialized by the same method.
 }
\label{tab_mlcec_error_rate}
\end{table}

\subsection{DCEC for face recognition}

In combination with the advantages of CEC and feature-level strategy, DCEC was employed to learn discriminative representation of human faces.

\begin{figure}[!htb]
    \centering
    \includegraphics[width=0.07\textwidth]{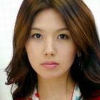}\hspace{0.3em}
    \includegraphics[width=0.07\textwidth]{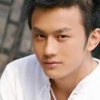}\hspace{0.3em}
    \includegraphics[width=0.07\textwidth]{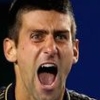}\hspace{0.3em}
    \includegraphics[width=0.07\textwidth]{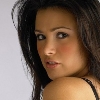}
    \caption{Examples of the Webface database. Through large range of age, expression, pose, and external environment, the database contains eastern and western people of quasi-equal number.}
    \label{fig_example_webface}
\end{figure}

\textbf{Data.} For training our DCEC, we collected a set of face images from websites, and formed a database called \emph{Webface} (Fig.~\ref{fig_example_webface}). It contains 156,398 face images of 4,024 identities, most of which are celebrities. Each identity owns quasi-equal number of images. All these images were roughly aligned according to a group of landmarks \cite{yan2013learn}, and normalized to the size of $100\times100$ with RGB channels. Finally, 3,500 identities were selected to form the training set, and the rest were devoted to the validation. We tested our model on LFW \cite{huang2007labeled} with their official unrestricted protocol. The identities of Webface and LFW are exclusive.

\textbf{Setting.}
To build the CNN module, we adopted one convolutional layer and two locally-connected layers, each of which was followed by a max-pooling layer. Locally-connected layer is similar to convolutional layer, while it does not share weights within feature maps. Therefore, it is suitable to extract features from a set of regular images, \eg human faces. As to CEC, the encoder and the decoder were both of single FC layer. The network employed ReLU as activation function. The softmax corresponded to the training identities. See Table.~\ref{tab_dcm_architecture} for the details of the parameters.

\begin{table} \small
 \newcommand{\tabincell}[2]{\begin{tabular}{@{}#1@{}}#2\end{tabular}}
 \tabcolsep 5
 pt
  \begin{center}
  \begin{tabular}{| c | c | c | c |}
    \hline
    \tabincell{c}{Name} & Type & \tabincell{c}{Filter Size/ \\ Stride} & \tabincell{c}{Output \\Size}  \\
    \hline

    Conv1 & conv. & $3\times3/1$ & \tabincell{c}{ $100\times100$ $\times32$} \\

    \hline
    Pool1 & \tabincell{c}{max pooling}  & $2\times2/2$ & \tabincell{c}{$50\times50$ $\times32$} \\

    \hline
    Local2 & \tabincell{c}{local}  & $3\times3/1$ & \tabincell{c}{$50\times50$ $\times64$} \\

    \hline
    Pool2 & \tabincell{c}{max pooling} & $2\times2/2$ & \tabincell{c}{$25\times25$ $\times64$} \\

    \hline
    Local3 & \tabincell{c}{local}  & $3\times3/1$ & \tabincell{c}{$25\times25$ $\times128$} \\

    \hline
    Pool3 & \tabincell{c}{max pooling}  & $2\times2/2$ & \tabincell{c}{$13\times13$ $\times128$} \\

    \hline
    $h_1$ & \tabincell{c}{FC}  &   N/A   & 512 \\

    \hline
    $h_2$ (encoder) & \tabincell{c}{FC}  &    N/A   & 256 \\

    \hline
    $z$ (decoder) & \tabincell{c}{FC}  &    N/A  & 512 \\

    \hline
    Softmax & \tabincell{c}{softmax}  &  N/A  & 3500 \\
    \hline

  \end{tabular}
  \end{center}

  \caption{Parameters of the architecture of DCEC for face representation learning. Both the layer $z$ (decoder) and softmax followed the layer $h_2$ (encoder).}

  \label{tab_dcm_architecture}
\end{table}

Each image was horizontally flipped to double the data amount. We generated totally about 25 million intra-person pairs.
The CNN module and the CEC were trained together, according to the objective (Eq.~\ref{dcec_objective}).

After training, we extracted the feature $h_2$, which was then processed by PCA and Joint Bayesian (JB) \cite{chen2012bayesian} for face verification.
We implemented the test under the LFW official unrestricted protocol.
Besides, recent studies \cite{liao2014benchmark} have noticed the limitations of the original LFW evaluation, \eg, limited pairs for verification, high FAR, and no identification experiments.
Therefore, we also tried the BLUFR protocol proposed in \cite{liao2014benchmark}, which included both verification and open-set identification experiments with an exhaustive evaluation of over 40 million of matching scores.

\textbf{Result.}
We compared our DCEC with the network that trained by only softmax. We also compared it with contrastive-style DeepID2 and DeepID2+ \cite{sun2014deep,sun2014deeply}, which used the similar structure (softmax + contrastive cost).

It should be noted that, though increasing higher results have been reported on LFW, it is not clear about the influence of the large private training data they used.
To make a fair comparison, we trained all the networks on the same Webface database, respectively.

The results are listed in Table.~\ref{tab_dcm_compare_deepID2}. Our DCEC yielded the best results under all the protocols.
The \emph{softmax-only} column shows that the absence of class-encoder leads to significant depravity of performance.
Hence, the improvement of DCEC was mainly attributed to the class-encoder.

The BLUFR evaluation indicated that the proposed method performed better under practical scenarios like verification at low FARs and the watch-list task in surveillance.

\begin{table} \small
 \newcommand{\tabincell}[2]{\begin{tabular}{@{}#1@{}}#2\end{tabular}}
 \tabcolsep 5
 pt
  \begin{center}
  \begin{tabular}{| c | c | c | c | c |}
    \hline
    & DeepID2   &   DeepID2+    &   \tabincell{c}{Softmax \\ only}    & DCEC \\
    \hline
    \hline

    \tabincell{c}{VR (\%) \\ PCA+JB} & 94.97 & 95.33    & 94.21 & \textbf{95.87} \\
    \hline

    \tabincell{c}{VR (\%) \\ @FAR=0.1\%} & 55.51 & 57.13    &   38.61 & \textbf{57.22} \\
    \hline

    \tabincell{c}{DIR (\%) \\ @FAR=1\%, \\ Rank=1} & 20.19 & 15.27 & 12.38 & \textbf{21.58} \\
    \hline

  \end{tabular}
  \end{center}

  \caption{The first line shows the accuracies under LFW unrestricted protocol. The second and the bottom lines indicate the two criteria of the BLUFR protocol, respectively.}

  \label{tab_dcm_compare_deepID2}
\end{table}

To eliminate the background, we cropped the face images according to 7 patches used in Sun \etal \cite{sun2014deep}, and trained 7 DCECs with them.
We fused the 7 models and tested them on the YouTube Faces (YTF) database \cite{wolf2011face}. This gave a competitive performance (Table.~\ref{tab_dcm_compare_ytf}). Note that DeepFace \cite{taigman2014deepface} used much more data (4.4 million images) and deeper architecture than ours.

\begin{table} \small
 \newcommand{\tabincell}[2]{\begin{tabular}{@{}#1@{}}#2\end{tabular}}
 \tabcolsep 5
 pt
  \begin{center}
  \begin{tabular}{| c | c |}
    \hline
    Method   &   VR (\%)     \\
    \hline
    \hline

    LM3L \cite{hu2014large}  & 81.3 $\pm 1.2$  \\
    \hline

    DDML (LBP) \cite{hu2014discriminative}  & 81.3 $\pm 1.6$   \\
    \hline

    DDML (combined) \cite{hu2014discriminative}  & 82.3 $\pm$ 1.5  \\
    \hline

    EigenPEP \cite{li2014eigen}  & 84.8 $\pm$ 1.4  \\
    \hline

    DeepFace-single \cite{taigman2014deepface}  & \textbf{91.4 $\pm$ 1.1}   \\
    \hline

    DCEC (fusion)  & \textbf{90.2 $\pm$ 0.4}   \\
    \hline

  \end{tabular}
  \end{center}

  \caption{Comparison on the YTF database, with the first two accuracies in bold.}

  \label{tab_dcm_compare_ytf}
\end{table}

\textbf{Analysis.}
Our DCEC used only intra-class pairs for training, and obtained better results than DeepID2 and DeepID2+ which used both intra- and inter-class pairs.
It implies that inter-class pairs contribute very little for training.
In addition, rather than the penalty by feature distance (contrastive cost), intra-class reconstruction gives better regularization for learning robust and discriminative face representation.
There are two reasons for this. First, the $L_2$ contrastive cost gives limited effect in the high-dimensional feature space, whereas the class-encoder minimizes the intra-class variation implicitly.
Second, in the high-dimensional space, the discriminative methods often allocate much larger partition than the proper class, leading to false positives with high confidence \cite{nguyen2014deep}.
By contrast, the generative method, involved in CEC, eliminates the nuisance factors in the feature space with their low marginal probability.

\textbf{Negative pairs.}
DCEC does not require inter-class pairs (the negatives). This can accelerate the training process comparing with the contrastive-style methods or the margin-style methods (often with time-consuming hard-negative-mining).

\section{Conclusion}
In this paper, we have two main contributions.

Firstly, we propose a novel class-encoder model, which minimizes the intra-class variations and learns discriminative manifolds of data at a class scale.
The experiment on MNIST shows that, if data is well aligned and with mono-background, the mere data-level reconstruction is able to bring about discrimination in not only the decoder, but the encoder as well.
We further imposed the class-encoder into the softmax classifier and improves the ability of generalization. The intra-class convergency leads to a sharp priori distribution, from which we obtain high value of conditional probability to the correct prediction given the trained weight matrix and the inputs.

\begin{figure}[!htb]
    \centering
    \includegraphics[width=0.135\textwidth]{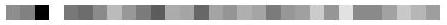}\hspace{1em}
    \includegraphics[width=0.135\textwidth]{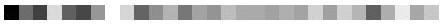}\hspace{1em}
    \includegraphics[width=0.135\textwidth]{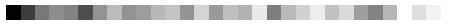}\\  \vspace{0.2em}
    \includegraphics[width=0.08\textwidth]{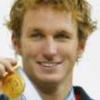}\hspace{3em}
    \includegraphics[width=0.08\textwidth]{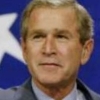}\hspace{3em}
    \includegraphics[width=0.08\textwidth]{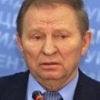}\\

    \includegraphics[width=0.135\textwidth]{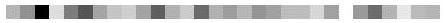}\hspace{1em}
    \includegraphics[width=0.135\textwidth]{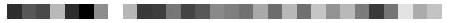}\hspace{1em}
    \includegraphics[width=0.135\textwidth]{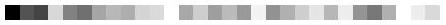}\\  \vspace{0.2em}
    \includegraphics[width=0.08\textwidth]{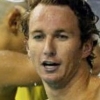}\hspace{3em}
    \includegraphics[width=0.08\textwidth]{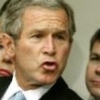}\hspace{3em}
    \includegraphics[width=0.08\textwidth]{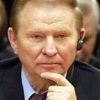}\\

    \caption{Instances in LFW and the corresponding feature vectors learned by DCEC. Each column belongs to an identity. 
    }
    \label{fig_lfw_image_feature}
\end{figure}

Secondly, we generalize the class-encoder to the feature-level, and combine the convolutional network and the CEC to learn discriminative features (Fig.~\ref{fig_lfw_image_feature}). Our DCEC obtained competitive results with much less training data regarding to state-of-the-art on face recognition. The feature-level strategy has well coped with size issue and translation issue of FC networks; and CNNs have gained better generalization from class-encoder.

\section{Acknowledgments}

This work was supported by the Chinese National Natural Science Foundation Projects \#61375037, \#61473291, \#61572501, \#61502491, \#61572536, National Science and Technology Support Program Project \#2013BAK02B01, Chinese Academy of Sciences Project No.KGZD-EW-102-2, and AuthenMetric R\&D Funds. The Tesla K40 used for this research was donated by the NVIDIA Corporation.

{\small
\bibliographystyle{ieee}
\bibliography{cecr}
}

\end{document}